\documentclass{article}

    \usepackage[preprint]{neurips_2020}

\usepackage[utf8]{inputenc} 
\usepackage[T1]{fontenc}    
\usepackage{amsmath,amsfonts,amsthm} 

\usepackage{hyperref}       

\newtheorem{theorem}{Theorem}

\newtheorem{property}{Property}
\usepackage{multirow,bigdelim}
\usepackage{graphicx} 

\usepackage{caption}
\usepackage{subcaption}
\usepackage{url}            
\usepackage{booktabs}       
\usepackage{amsfonts}       
\usepackage{nicefrac}       
\usepackage{microtype}      
\usepackage{xcolor}

\title{What needles do sparse neural networks find in nonlinear haystacks}

\author{ Sylvain Sardy\\ Department of Mathematics\\ University of Geneva\\ \texttt{sylvain.sardy@unige.ch} \\ 
  \And Nick Hengartner \\ Los Alamos National Laboratory \\ Theoretical Biology and Biophysics group \\ \texttt{nickh@lanl.gov} \\
  \And Nikolai Bonenko  \\ Department of Mathematics\\ University of Geneva\\ \texttt{nikolai.bobenko@unige.ch}  \\
   \And Yen Ting Lin\thanks{Corresponding author}  \\ Los Alamos National Laboratory \\ Information Sciences group \\ \texttt{yentingl@lanl.gov} \\
}

\begin{document}

\maketitle

\begin{abstract}
  Using a sparsity inducing penalty in artificial neural networks (ANNs) avoids over-fitting, especially in situations where noise is high and the training set is small in comparison to the  number of features.  For linear models, such an approach provably also  recovers the important features with high probability in regimes for a well-chosen  penalty parameter.  The typical way of setting the penalty parameter is by splitting the data set and performing the cross-validation, which is (1) computationally expensive and (2) not desirable when the data set is already small to be further split (for example, whole-genome sequence data).  In this study, we establish the theoretical foundation to select the penalty parameter without cross-validation based on bounding with a high probability the infinite norm of the gradient of the loss function at zero under the zero-feature assumption. Our approach is a generalization of the universal threshold of Donoho and Johnstone (1994) to nonlinear ANN learning.  
  We perform a set of comprehensive Monte Carlo simulations on a simple model, and the numerical results show the effectiveness of the proposed approach.
\end{abstract}

\section{Introduction}

Machine Learning seeks to empirically extract rules from 
data to make predictions on future observations collected under similar
circumstances.  This learning can be supervised (regression or classification) or unsupervised (clustering). The quality of predictions on new data
determines how good the learned rules are, as in many applications,
obtaining reliable predictions is the end goal.  

Over the past ten years, Artificial Neural Networks (ANNs) have
become the model of choice for machine learning thanks  to 
the quality of the predictions they exhibit 
in many modern applications.  Their success, in part,
can be attributed to the expressiveness of ANNs.  However, 
traditional measures of model complexity based on the number of
parameters do not apply.   For one, ANNs are over parametrized with 
multiple distinct settings of the parameters
leading to the same prediction.  This makes understanding and
interpreting the predictions challenging.
Yet in scientific applications, one often seeks to do just that.

In keeping with Occam's razor, among all the models with similar
predictive capability, the one with the smallest number of features
should be selected.  Statistically, models with fewer features 
not only are easier
to interpret but can produce predictors with good statistical properties
because such models disregard useless features that contribute only to higher variance.

Operationally, the model selection paradigm often 
uses cross-validation in which the data is randomly split and 
models are built on a training set and predictions are evaluated 
on the testing set.  While conceptually elegant, cross-validation
is of limited use if fitting a single model is computationally expensive
(in that case, we can not train very many models) or the sample
size is small (in which case, splitting the data leaves few observations
to fit the model).  

Since ANNs and in particular deep ANNs are computationally expensive to fit, cross-validation is not often used to do feature selection.   
In addition, quadratic prediction error from cross-validation 
 exhibits an unexpected behavior with ANNs.  As expected,
the training  error always decreases with increasing number of input features.  
While the quadratic prediction error on the test set is at first U-shaped (initially decreasing thanks to  decreasing bias, and then increasing due to an excess of variance), it then unexpectedly decreases a second time.  This phenomenon known as \emph{double descent} has been empirically observed \citep{AdvaniSaxe2017,ClementHongler2019}.
For least squares estimation regularized by an $\ell_2$ ridge penalty \citep{ridgeHK}, double descent has been mathematically described for two-layer ANNs with random first-layer weights
by \citet{MeiMontanari2019} and \citet{HastieMRT2019}.
They show that for high signal-to-noise ratio (SNR) and large sample size, high complexity is optimal for the ridgeless limit estimator of the weights, leading to a smooth and more expressive interpolating learner. In other words, interpolation is good and leads to double descent, which after careful thinking should not be a surprise since the interpolating ANN becomes smoother with increasing number of layers, and therefore better interpolates between training data. Indeed with high SNR, the signal is almost noiseless, so a smooth interpolating function shall perform well for future prediction. But data are not always noiseless, and in noisy regimes, that is with low SNR and small sample size, \citet{MeiMontanari2019} also observe that regularization is needed, as expected.

\medskip 

In this paper, we present an alternative to cross-validation geared towards identifying important features.
Specifically, we develop an automatic feature screening
method for 
simultaneous features extraction and generalization. For ease of
exposition,  we present
our novel method in the context of regression, noting that the
ideas can be ported to classification and beyond.

Our approach exploits ideas from statistical hypothesis testing that
directly focus on identifying significant features, and this 
without explicitly considering
minimizing the generalization error.  Similar ideas percolate the statistics
literature, see for example \citet{JS04}, \citet{CDS99}, \citet{Tibs:regr:1996} with the lasso, \citet{BuhlGeer11} who
propose methods for finding {\em needles in a haystack} in linear models. 
In this context, the optimized criteria is not the prediction error,
but measures the ability of the algorithms to 
retrieve the needles (i.e., relevant features).  Useful criteria include the stringent exact support recovery criterion, and softer criteria such as
the false discovery rate (FDR) \citep{Benj:Hoch:1995}, true positive rate (TPR), and screening (i.e., including all relevant features).
Of course some regularization methods have already been developed to enforce sparsity to the weights of ANNs.   
We are not aware of any of these methods having been applied to feature selection in ANNs.

Methods leading to sparse ANNs  have already been developed.   For example, {\em dropout} leaves out a certain number of 
neurons to prevent overfitting, which incidentally can be used to perform 
feature selection \citep{DBLP:journals/corr/abs-1207-0580,DBLP:journals/jmlr/SrivastavaHKSS14}. 
That approach is computationally expensive, as it
is a combinatorial problem to decide which neurons in the ANN 
to leave out and which to keep in.  Sparse neuron architectures can be achieved by other means:
\citet{pmlr-v70-mollaysa17a} enforce sparsity based on the Jacobian
and \citet{10.55552976456.2976557,10.5555/2981562.2981711,MemoryBound2014,DBLP:journals/corr/abs-1901-01021} employ $\ell_1$-based penalty of lasso to induce sparsity.

All of these sparsity inducing methods suffer from two drawbacks: (1) the selection of the regularization parameter is rarely addressed, and when it is, the selection is based on the computer intensive cross-validation geared towards good generalization performance; (2) the ability to recover the ``right'' features has not been investigated based on support recovery, FDR, TPR and screening. 

\medskip
This paper is organized as follows.  In Section 2, we present the theoretical
framework of our ANN feature selection method.
It generalizes the universal threshold of \citet{Dono94b} 
to  the non-convex optimization setting.  
In Section 3, we evaluate via simulations, the ability of our method to 
recover the true features in the challenging regime of 
low SNR and small sample size, where interpolation is not 
appropriate and where regularization improves generalization.  
Specifically, 
assuming that only a few features are informative, 
we seek to recover a sparsely encoded ANN by 
adding an $\ell_1$-based sparsity inducing penalty
and a selection of the 
magnitude of the $\ell_1$ penalty based on the theory presented in Section~\ref{subsct:lambda},
and we evaluate the effectiveness of our method using the stringent exact support recovery criterion.

\section{Theoretical foundation}  \label{sct:tf}

\subsection{Regression model and notation}

Let $\{(y_i,x_i) \}_{i=1}^n$ be $n$ realizations 
from the pair of random variable and vector $(Y,X)$, 
for which the scalar response $Y \in {\mathbb R}^1$ is 
related to a $p_1\times 1$ real-valued feature vector $X$
through the conditional expectation 
\begin{equation} \label{eq:condexpect}
{\mathbb E}[Y\mid X=x] = \mu(x),
\end{equation}
for some unknown function $\mu$. We assume here that the noise on the output variable $Y$ is Gaussian $\mathcal{N}\left(0, \xi^2\right)$, where the variance $\xi^2$ is unknown.
We model $\mu$ as a standard fully connected ANN with $l$ layers 
\begin{equation}
\mu_{ \theta}(x)=  S_l \circ \ldots \circ S_1\left(x\right), 
\end{equation}
where, in each layer $k < l$, the nonlinear function $S_k(u)=\sigma(W_k  u +  b_k)$ maps the $p_k\times 1$ vector $u$ into a $p_{k+1}\times 1$ latent vector obtained by applying 
 an activation function $\sigma$ component-wise to $W_k  u +  b_k$, where
 $W_k$ is a $p_{k+1} \times p_k$ matrix of weights, 
$b_k$ is a $p_{k+1}\times 1$ vector of biases, and the operation $+$ is the broadcasting operation.
To predict a scalar value, the last function $S_l$ is a linear combination of the entries of the $p_{l-1}\times 1$ latent vector created at the previous layer plus an offset/intercept $b_l \in \mathbb{R}^1$.
The parameters indexing this neural network are therefore ${ \theta}=(W_1, b_1, \ldots,
W_l,  b_l)$. 

Our regression goal is two-fold. We want to generalize well, that is, given a new vector of features/input, we want to predict the output with precision. We also believe that only a few features in the $p_1$-long input vector carry information to predict the output. So our second goal is to find needles in the haystack by selecting a subset of the $p_1$-long input. 
For instance, $x$ can be a vector of thousands of gene expression, and genetic aims to identify the ones having an effect.

\subsection{Sparse estimation}

The needles finding goal is achieved in our model by the matrix of weights $W_1$ of which some entries are adaptively set to zero with our method. Identifying the features in $x$ corresponding to non-zero entries in $W_1$ amounts to finding the needles.
So for the first layer, the weights $W_1$ will be regularized in a way that induces sparsity. 
If the weights of the other layers are not constrained, it may create undesirable effects, such as unbounded weights and consequently unbounded learned dictionaries/latent variables, or redundant parametrization of the network with small and large weights compensating one another.
To alleviate issues of unbounded dictionaries and unidentifiability, we constrain the weights at any level $k > 1$ to be in the $\ell_2$-sphere of radius one in the following way. 
We define  the $j^\text{th}$ nonlinear function $S_{k,j}$  in layer $k$  as
\begin{equation}
S_{k,j}(u)= \left \{ \begin{array}{ll}
\frac{ \langle { w}_k^{(j)},u  \rangle}{\left \Vert { w}_k^{(j)} \right \Vert_2}+b_l & k=l \\
\sigma\left(\frac{ \langle { w}_k^{(j)}, u  \rangle}{\left \Vert{ w}_k^{(j)} \right\Vert_2}+b_{k,j}\right) & 1< k <l \\
\sigma\left( \langle { w}_k^{(j)}, u \rangle +b_{k,j}\right) & k=1 \end{array}
\right . , \quad j \in \{1,\ldots,p_{k+1}\},
\end{equation}
where ${ w}_k^{(j)}$ is the $j^\text{th}$ row of $W_k$.
We pose here to make an important remark about the biases $b_{k,j}$, which is the $j^\text{th}$ entry of the column vector $b_k$: in approximation theory, $\sigma$ can be seen as basis functions (e.g., splines or wavelets) translated by an amount $b_{k,j}$ chosen to fit the data within the range of the latent variables created at layer $k-1$. So we impose the following constraint on the biases.

\bigskip
{\bf Biases constraint}.  Given a $p_1 \times n$ input matrix $U_0$, which is composed by horizontally stacking $n$ input vectors (which are $p_1\times 1$ column vectors), and a sequence of weights $W_1, \ldots, W_l$ and  variables $U_k=[{\bf u}_{k,1}, \ldots, {\bf u}_{k,n}]=W_k U_{k-1}$ at each layer $k\in\{0,1,\ldots, l\}$, we impose the following constraint on the biases:
\begin{equation} \label{eq:biasescontraint}
b_{k,j} \in  \left \{  \begin{array}{ll}
\mathbb{R} & k=l, \\
\frac{1}{\|{ w}_k^{(j)} \|_2} \left [ \min_{i=1}^n  \langle {w}_k^{(j)}, { u}_{k-1,i}  \rangle, \max_{i=1}^n \langle { w}_k^{(j)}, { u}_{k-1,i}  \rangle \right ] & 1< k < l , \quad j \in \{1,\ldots,p_{k+1}\}\\
\left [ \min_{i=1}^n  \langle { w}_k^{(j)}, { u}_{k-1,i}   \rangle, \max_{i=1}^n  \langle { w}_k^{(j)}, { u}_{k-1,i}   \rangle \right ] & k = 1.
\end{array} \right .
\end{equation}
\bigskip

At the last layer $k=l$, the scalar bias $b_l\in \mathbb{R} $  plays the role of an \emph{unconstrained} offset/intercept.

Sparsity in the first layer allows interpretability of the fitted model.
We enforce sparsity and  control overfitting by minimizing a compromise between a measure~$l$ of closeness
to the data and a measure of sparsity. Letting ${ y}$ be the vector of all training responses and ${\mu}_{ \theta}({ x})$ be their predicted values at all training locations ${ x}$, we estimate the parameters ${ \theta}$ of the ANN by choosing the best local minimum found by a numerical scheme to
\begin{equation} \label{eq:L1}
\hat { \theta}_\lambda = \arg \min_{ \theta\in {\Theta}} l({y} , {\mu}_{\theta}( x)) + \lambda \|W_{1}\|_1,
\end{equation}
where $\lambda>0$ is the regularization parameter.

The $\ell_1$-penalty as a mean to induce sparsity is reminiscent of waveshrink~\citep{Dono94b} and lasso~\citep{Tibs:regr:1996}, and has been considered in neural networks (see for instance \citet{DeepFeatureSelection2016,DBLP:journals/corr/abs-1901-01021}). For linear associations $\mu_{  \theta}({x})=\sum_{i=1}^{p_1} \theta_i x_i$ with a sparse vector ${  \theta}$, the lasso has the remarkable property of retrieving the non-zero entries of ${  \theta}$ in certain regimes (that depend on $n$, $p_1$, SNR, training locations ${\bf x}$ and amount of sparsity); this has been well studied \citep{CandesTao05,DonohoDL06,6034731,BuhlGeer11}. Our contribution is to investigate whether this property extends to nonlinear associations with ANNs to discover their underlying lower-dimensional structures.
We propose  specific goodness-of-fit measure $l$, activation function $\sigma$, bias constraints~\eqref{eq:biasescontraint}  along with an efficient and pertinent selection of the regularization parameter~$\lambda$.

Our $\lambda$ hinges on retrieving the constant function with high probability for nonlinear ANNs.
For linear models  in wavelet denoising theory  \citep{Dono94b}, this approach led to an asymptotic minimax property to retrieve a function and its sparse wavelet representation in Besov spaces. So we ask our activation function to have the following property.

\bigskip

{\bf Activation function requirement}. The activation function $\sigma$ must be unbounded. Moreover it must be null and have a positive derivative at zero:
\begin{equation} \label{sigma(0)=0}
\sigma(0)=0 \quad {\rm and} \quad \sigma'(0)>0.
\end{equation}
A possible function satisfying the requirements is $\sigma(u)=\log(1+\exp(u))-\log 2$.

Given the biases constant \eqref{eq:biasescontraint}  and the activation function requirement \eqref{sigma(0)=0}, we have the following property for which the proof is immediate.
\begin{property} 
 \label{eq:propconstant}
Assuming  \eqref{eq:biasescontraint}  and  \eqref{sigma(0)=0}, then setting the first layer weights $W_{1}$ to the zero matrix implies $\mu_{ \theta}({ x})=b_l$ is constant for all~${x} \in \mathbb{R}^{p_1}$. 
\end{property}

\subsection{Selection of regularization parameter $\lambda$}
 \label{subsct:lambda}

The choice of $ \lambda$ is based on Property~\ref{eq:propconstant}. The quantile universal threshold \citep{Dono94b, Dono95asym,Giacoetal17} aims at retrieving the constant function with high probability by setting all parameters to zero. 
The quantile universal threshold has so far been developed and employed for cost functions that are convex in the parameters, hence guaranteeing that any local minimum is also global. For the cost function in~\ref{eq:L1} that is not convex in the parameters we extend the quantile universal threshold to guarantee a local minimum at the sparse point of interest $W_{1}=O$  that is the null matrix of weights.
Since the term $\lambda \|W_{1}\|_1$ is part of the cost function in~\eqref{eq:L1}, then we seek $\lambda>0$ such that with high probability $\hat \theta=(\hat W_1 ,\hat { b}_1, \ldots, \hat W_l ,\hat  b_l)$ is a local minimum to~\eqref{eq:L1} with $\hat W_{1}=O$, leading to constant prediction by Property~\ref{eq:propconstant}.
 
 \begin{theorem}
 \label{thm:lambda_choice}
  Consider the optimization problem~\eqref{eq:L1} and define ${ g}_0({ y}, { x})=\nabla_{W_{1}} l({ y},{\mu}_{ \theta}({ x}))$ evaluated at
 the null matrix for $W_{1}$ (hence  ${ b}_{1}={0}$ and all higher layer return null values), at the sample average $\hat b_l=\bar{ y} := \frac{1}{n} \sum_{i=1}^n y_i$
  and at any value $W_{k}$ and ${b}_k$ for the other layers.
  If $\lambda > \sup_{(W_2, b_2, \dots, b_{l-1}, W_l)}\| { g}_0({ y}, {x}) \|_\infty$ and $l$ is sufficiently smooth, then there is a local minimum  to~\eqref{eq:L1} at $\hat W_{1}$ set to the null $p_2 \times p_1$ matrix.
 \end{theorem}

 The proof of Theorem \ref{thm:lambda_choice} is provided in the supplementary material. 

 Note that the choice of $\hat b_l = \bar{ y}$ is justified by observing that due to Property \ref{eq:propconstant} at $W_1 = 0$ and for the loss function $l(\cdot) = ||\cdot||_2$ we are in the setting of a quadratic form which is minimized for  $\hat b_l=\bar y$. 
 
 \begin{theorem} Consider a training set $({ y}, {x})$.
 Define the random vector $G_0={ g}_0({ Y}, { x})$ where ${ Y}$ is random vector simulated under the null hypothesis $H_0: W_{1}=O, {b}_{1}={ 0}$, that is $H_0: \mu_{ \theta}=b_l$ is the constant function.
  Let $\Lambda=\| { G}_0 \|_\infty$ and define the quantile universal threshold $\lambda_{\rm QUT}=F^{-1}_\Lambda(1-\alpha)$ for a small value of $\alpha$.
Then,
\begin{equation}
{\mathbb P}_{H_0}(\mbox{there exists a local minimum to~\eqref{eq:L1} such that } \mu_{\hat { \theta}_{\lambda_{\rm QUT}}}={\rm constant})=1-\alpha.
\end{equation}
 \end{theorem}
 
The law of $\Lambda$ is unknown but can be estimated by Monte Carlo simulation, provided it does not depend on the remaining parameters (the last layer bias $b_l$ and the parameters of the noise measurements) of the fully sparse neural network under $H_0$. Inspired by square-root lasso \citep{BCW11},  the following theorem states that for Gaussian noise errors with unknown variance $\xi^2$ and for square-root $\ell_2$-loss $l$ in~\eqref{eq:L1}, the statistic $\Lambda$ does not depend on any unknown parameter.
This would not be true for the  square $\ell_2$-norm since the law of $\Lambda$ would depend on $\xi$, which is hard to estimate in high-dimension. So using the $\ell_2$-norm as loss function alleviates a difficult variance estimation problem.
 \begin{theorem}
  \label{thm:pivot}
Assuming the conditional expectation in~\eqref{eq:condexpect} is based on Gaussian errors with unknown variance, then choosing the loss $l({ y} , {\mu}_{\boldsymbol \theta}( x))  =\| { y} -  \mu_{ \theta}({x}) \|_2$ makes the statistic $\Lambda$  pivotal (that is, not a function of any parameter, including the unknown noise variance).
 \end{theorem}
 
The proof of Theorem \ref{thm:pivot}  stems from the fact that the gradient of the square root of the $\ell_2$-loss has a numerator and denominator that are proportional to $\xi$ and have responses $y^{(k)}$ centered around $\bar y$. Hence the gradient depends neither on  $\xi$ nor on $b_l$ (see for instance the formula of the gradient for a two layer neural network in equation (1) of the supplementary material)
 
Once the quantile universal threshold $\lambda_{\rm QUT}$ is calculated, we solve~\eqref{eq:L1} first by steepest descent with a small learning rate, and then employ a proximal method to refine the minimum with a more sparse solution that exactly sets to zero some entries  of $\hat {\boldsymbol \theta}_{\lambda_{\rm QUT}}$ \citep{FISTA09,10.1561/2200000015}.

 \section{Simulation study}

As in~\citet{MeiMontanari2019}, we consider a two-layer ANN by assuming that the underlying association is a sparse ANN, that is
$$
\mu_\theta(x)=b_2+ \frac{1}{\sqrt{h}} \sum_{i=1}^h \sigma(x_{2i}-x_{2i-1})=10+ \frac{ \langle { w}_2, {u} \rangle}{\left \Vert{ w}_2\right \Vert_2}  
$$
with ${ w}_2=({ 1}_h$, ${ 0}_{p_2-h})$, $b_2=10$,
${ u}=\sigma(W_1 \cdot x +b_1)$, $b_{1,j}=0$ and 
\newcommand\coolrightbrace[2]{%
\left.\vphantom{\begin{matrix} #1 \end{matrix}}\right\}#2}
\begin{equation}
W_1 =\left [ \begin{array}{ccccccccccccc}
-1 & 1 & 0 &  0 &  \ldots &  & & &  & &\ldots & 0\\
 0 & 0 & -1 &  1 &  0 &  \ldots & & & & &  \ldots& 0\\
    &&&& \vdots &\vdots&&& && &  \vdots \\
0 & \ldots &  & & \ldots&  0 & -1 & 1 & 0 &  \ldots & \ldots &  0 \\
0 & \ldots & &&& \ldots &&& & &\ldots & 0 \\
\vdots &  &&&&&&&&&& \vdots \\ 
0 & \ldots &  &&& \ldots &&&&&\ldots& 0
\end{array}
\right ]
\begin{matrix}
    \coolrightbrace{0 \\ 0 \\ \vdots \\ 0 }{h\hspace{21.5pt}}\\
    \coolrightbrace{0 \\ \vdots \\ 0 }{p_2-h}
\end{matrix}
\end{equation}
for $\log_2 h \in \{0,1,2 \}$. This corresponds to 2, 4 and 8 needles in a nonlinear haystack of size $p_1$ with $\log_2 p_1 \in \left\{4,5,6,7,8,9\right\}$

Based on the information in a training set, our main goal is to recover the sparse structure of $W_1$, that is, not the exact values of $-1$ and $+1$ but the exact location of the non-zero values in $W_1$.
We are interested in low signal-to-noise ratio so we consider a training set of only $n=300$ samples independently drawn from a standard multivariate Gaussian leading to $\{x_i\}_{i=1}^n$ in ${\mathbb R}^{p_1}$ and the 
noisy measurements $\{y_i\}_{i=1}^n$ of $\mu$ at $x_i$
according to $y_i=\mu(x_i)+ \epsilon_i$  with $\mu=\mu_\theta$ by adding i.i.d.~standard Gaussian noise $\epsilon_i \sim   \mathcal{N}\left(0, \xi^2\right)$ with $\xi=0.1$ for $i=1,\ldots,n$. 
The total number of neurons used  in $\mu_\theta$ is $p_2$ with $\log_2 p_2 \in \left\{4,5,6,7,8,9\right\}$ while the number of active neurons is $h \in \{1,2,4\}$.

\begin{figure}
\includegraphics[width=\textwidth]{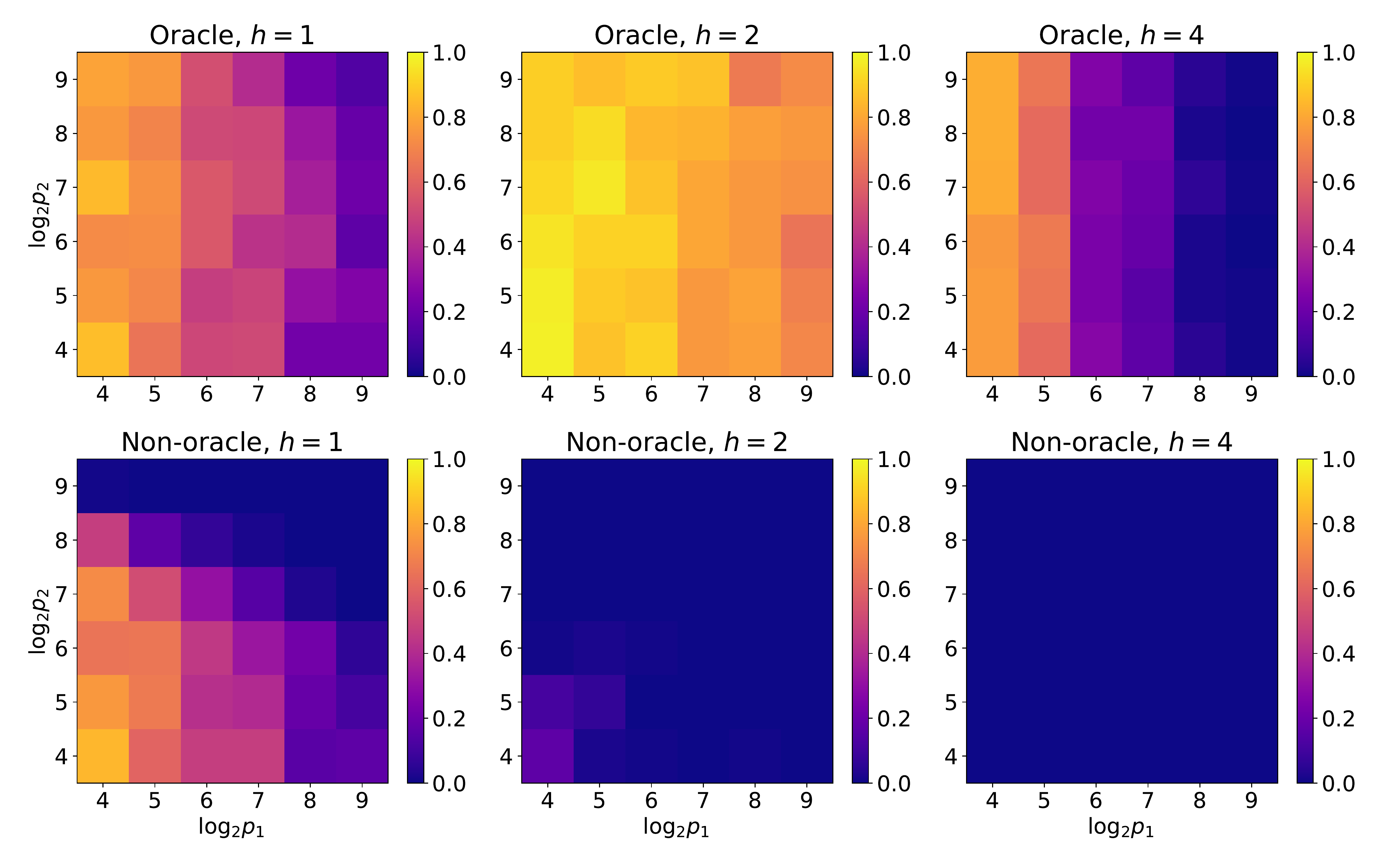}
\caption{Heatmap of estimated probability of support recovery based with the $\ell_1$-sparsity inducing penalty and the quantile universal threshold $\lambda_{\rm QUT}$ at $\alpha=0.05$ as a function of $(p_1,p_2)$ and increasing network complexity $h$ (from left to right). Top: oracle. Bottom: non-oracle.}
\label{fig:test1}
\end{figure}

To measure the ability of recovering the sparsity structure of $W_1$ we consider a stringent criterion that is difficult to achieve, even for the linear model:  the exact recovery of the sparsity of $W_1$. To that aim, for each scenario $(h,p_1,p_2)$, we simulate $M=100$ training samples of size $n=300$, calculate the corresponding $\lambda_{\rm QUT}$ and  parameter estimates $\hat {\boldsymbol \theta}_{\lambda_{\rm QUT}}$ by solving~\eqref{eq:L1}, extract $\hat W_1$ and evaluate the proportion of times (out of one hundred)  it matches the sparsity of the true $W_1$ that generated the training sample. To solve~\eqref{eq:L1}, we employed two strategies:
\begin{enumerate}
\item Oracle optimization: we initialize the optimization algorithm with the true $W_1$
\item Non-oracle optimization:  we start at a single random initial values.
\end{enumerate}
Optimists will look at the first option, but ANNs practitioner will consider the second.  Selecting the best outcome using multiple random restarts will move the reported performance from option 2 to 1.

  Figure~\ref{fig:test1} reports the estimated probabilities as a function of $(p_1,p_2)$ for $h\in2^{\{0,1,2\}}$ for the oracle strategy (top) and non-oracle strategy (bottom).
As for linear models, we observe there is a regime where we can retrieve $W_1$ with high probability. The goal of improving the non-oracle optimization is for the bottom plots to get closer to the top plots without the knowledge of the oracle. We also considered smaller and larger multiples of~$\lambda_{\rm QUT}$, but observed the results were not as good, showing that our choice of regularization parameter $\lambda$ is near optimal for exact needle/support recovery, at least in the regimes we considered.

\begin{figure}
\includegraphics[width=\textwidth]{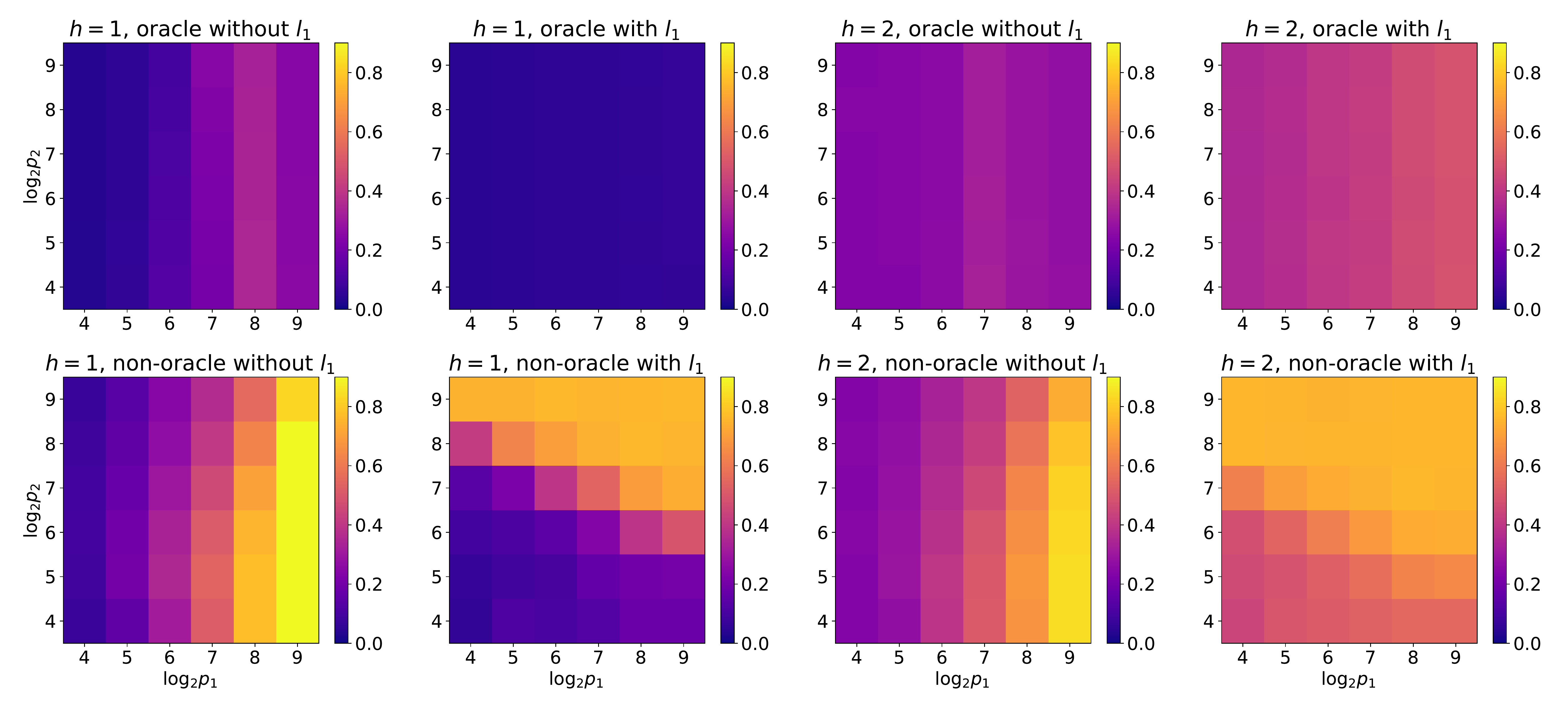}
\caption{Generalization as a function of $(p_1,p_2)$ with oracle (top) and non-oracle (bottom) strategies for $h=1$ (left four) and $h=2$ (right four) active neurons. For each 2x2 plots, the left plots show the results without regularization and the right plots with. Generalization on a test set is measured by the square root of the average $\ell_2$ loss between our predicted values and the true values at $30000$ new random locations.} 
\label{fig:test2}
\end{figure}

We also considered generalization:
for each case with predicted at a large number (here a hundred times the size $n$ of the training set) of new locations and reported the square root of the average $\ell_2$ loss between our predicted values and the true values.
Figure~\ref{fig:test2}  plots the results on the same scale. We observe that: (1) $\ell_1$ regularization with the quantile universal threshold $\lambda_{\rm QUT}$ outperforms no regularization in these regimes; (2) being oracle for the starting values of the parameters improves generalization, especially without regularization.
The conclusion is that regularization improves generalization when the size $p_1$ of the haystack is large, as long as the the complexity measured by$h$ is not too large. Otherwise the results are mitigated in the low SNR setting we are considering here. Larger training set and lower noise would allow to retrieve the sparsity structure of more complex ANNs. This shows that our method leads to the remarkable result that a sparse ANN  can generalize at least as well as a dense ANN in the settings we have considered. This calls for further developments of our method including: improving optimization for a cost function with a non-differentiable penalty, improving generalization by refitting the sparse ANN without a penalty to avoid shrinking the parameters  towards zero, along with deriving theoretical results to understand the regimes indexed by $(h,p_1,p_2,\xi )$ where our approach finds the needles.

 \section{Application}
 
 We consider genetic data measuring  the  expression  levels  of $p_1=  4088$  genes on $n=  71$  Bacillus  subtilis  bacteria  \citep{PeterBulbiology:14}.   The logarithms of gene expression measurements are known to have some strongly correlated genes, which also makes selection difficult.  The output is the riboflavin production rate of the bacteria. This is a high-dimensional setting in the sense that the training set is very small ($n=  71$) compared to the size of the haystack ($p_1=  4088$). Generalization is not the goal here, but finding the needles; the scientific questions are: what genes affect the riboflavin production rate? Is the association linear or not?
 
These data have  previously been employed to illustrate the property of the lasso to select needles in a linear model. The ground truth is not known here. Lasso-zero, a conservative method with low false discovery rate \citep{DesclouxSardy2018}, selects two genes whose indices are  $4003$ and $2564$. A less conservative version of lasso (based on the {\tt cv.glmnet} function in the {\tt glmnet} library of the R software) selects 30 needles including $4003$ and $2564$.

After calculating the quantile universal threshold $\lambda_{\rm QUT}=1.965$  for $p_2=8$ neurons and solving ~\eqref{eq:L1} with one hundred multiple starts, our approach finds a single neuron model with 30 needles, essentially the same ones as with the linear model. So the answers to the scientific questions are that at most 30 genes seem to be responsible for the riboflavin production rate of the bacteria and that a linear model seems sufficient since the selected number of neuron is one.

\section{Conclusion}

We demonstrated that, when noise is present and the training set is not extensive, $\ell_1$-regularization with our  specificities on the bias and activation function and with our prescribed selection of the penalty parameter not only has good generalization performances, but also can retrieve a sparse structure and identify pertinent features.
Our empirical results call for more theory to mathematically predict the regimes indexed by  $(h,p_1,p_2,\xi )$ where feature recovery is highly probable.

\section*{Broader Impact}
DNNs are widely used state-of-the-art black boxes.  There is a keen interest, especially in scientific and medical applications, to understand the “why” of model predictions.  Sparse encoding—automatic feature selection—provides a path towards such an understanding. The impediment of applying standard techniques developed in linear LASSO optimization for DNNs is the computational overhead required to estimate the magnitude of the penalty via cross-validation. Cross validation is neither desired when the sample size is already small and cannot be further partitioned.  Our work resolves these issues and makes sparse encoding closer to practical applications. The nature of our work is theoretical, and we do not envision potential negative impact to our society. 

\bibliographystyle{plainnat}

\end{document}


\maketitle

\section{Proof of Theorem 1 and a closed form expression}
\begin{proof}
	Let $\tilde{W}_1$ be any matrix satisfying $\norm{\tilde{W}_1}_1 = 1$ and denote by $\tilde{\theta_\epsilon}$ the parameters in ${ \theta}=(W_1, b_1, \ldots,
W_l,  b_l)$. 
 but with $W_1$ replaced with $\epsilon \tilde{W}_1$. Recall that we evaluate at $W_1 = 0$, which consequently sets $b_1=0$ given the biases constraint we require.

	Applying Taylor's theorem we then have
	\begin{align*}
		|l(y, \mu_{\tilde{\theta}_\epsilon}(x)) - l(y, \mu_\theta(x))| &= \nabla_{W_1} l(y, \mu_\theta(x)) (\epsilon \tilde{W}_1) + o(\epsilon^2 \norm{\tilde{W}_1}_1) \\
		&= \epsilon g_0(y,x) \tilde{W}_1 + o(\epsilon^2) \\
		&\leq \epsilon \norm{g_0(y,x)}_\infty + o(\epsilon^2).
	\end{align*}
	Therefore we get
	\[
		l(y, \mu_{\tilde{\theta}_\epsilon}(x)) \geq l(y, \mu_{\theta_\epsilon}(x)) - \epsilon \norm{g_0(y,x)}_\infty + o(\epsilon^2).
	\]
	Now in this calculation there is an implicit dependency on the other layers that are not fixed $\theta_r = (W_2, b_2, \ldots, b_{l-1}, W_l)$. Since we are looking for a universal lambda we need to take the supremum over all valid choices of these layers. Note that we assume normalization of all these layers and thus the supremum is finite. If we now assume that $\br{\lambda - \sup_{\theta_r}\norm{g_0(y,x)}_\infty} > C > 0$ it follows for the regularized loss function that
	\begin{align*}
		l(y, \mu_{\tilde{\theta}_\epsilon}(x)) + \lambda \norm{\epsilon \tilde{W}_1}_1 &\geq l(y, \mu_{\theta_\epsilon}(x)) + \epsilon \br{\lambda - \norm{g_0(y,x)}_\infty} + o(\epsilon^2) \\
		&> l(y, \mu_\theta(x)) + \epsilon C + o(\epsilon^2) \\
		&> l(y, \mu_\theta(x)) \quad \text{for $\epsilon$ small enough.}
	\end{align*}
	Thus the regularized loss functional with our choice of $\lambda$ indeed has a local minimum at $\theta$.
\end{proof}

We perform the explicit calculation of the gradient under square-root $\ell_2$-loss for a two layer neural network.
\begin{proof}
	We denote by $(y_k, x_k)$ the $k$-th training sample. As described above we have 
	\[
	l(y, \mu_\theta(x)) = \left(\sum_{k=1}^n \left\{y_k - (w_2\sigma(W_1x_k + b_1) + b_2)\right\}^2 \right)^\frac{1}{2}.
	\]
	When $W_1$ is set to the null matrix, then $b_1$ is constrained to be the null vector. Hence with the property that $\sigma(0)=0$, the least squares problem is solved for $b_2 = \bar{y}$. So we want to evaluate the gradient with respect to $W_1$ at $W_1 = 0, b_1 = 0, b_2 = \bar{y}$ which we call condition $(L_0)$. Let us consider the partial derivative of the $(i,j)$-th entry. Some elementary calculation yields
	\begin{align} \label{eq:Wij}
		\frac{\partial l(y, \mu_\theta(x))}{\partial W_{1;i,j}}\Big|_{(L_0)} &= \frac{\sigma'(0)}{\norm{y-\bar{y}}_2} \br{\sum_{k=1}^n (y_k -\bar{y}) w_{2;i} x_{k,j}} .
	\end{align}
	From Theorem 1, we need to find $\sup_{\norm{w_2}_2 = 1} \norm{\nabla_{W_1}l(y, \mu_\theta(x)) }_{\infty}$.
	Clearly 
	\begin{align*}
		\norm{\nabla_{W_1}l(y, \mu_\theta(x)) }_{\infty} &= \frac{\sigma'(0)}{\norm{y-\bar{y}}_2} \max_{i,j} \abs{\sum_{k=1}^n (y_k - \bar{y}) w_{2;i} x_{k,j}} \\
		&= \frac{\sigma'(0)}{\norm{y-\bar{y}}_2}  \|w_2\|_\infty \max_{j} \abs{\sum_{k=1}^n (y_k - \bar{y}) x_{k,j}}.
	\end{align*}
	With the contraint $\norm{w_2}_2 = 1$, we have that $\|w_2\|_\infty$ reaches a maximal value of one for any canonical vector. Thus we have
	\begin{align*}
		\sup_{\norm{w_2}_2 = 1} \norm{\nabla_{W_1}l(y, \mu_\theta(x)) }_{\infty} &= \frac{\sigma'(0)}{\norm{y-\bar{y}}_2} \norm{\sum_{k=1}^n (y_k - \bar{y}) x_k}_\infty.
	\end{align*}
		This is the closed form expression for $\norm{g_0(y,x)}_\infty$ of Theorem~1 for a two-layer ANN.
\end{proof}